\documentclass[journal]{IEEEtran}

\usepackage{times}
\usepackage{xspace}
\usepackage{amsmath, amsfonts}
\usepackage{array}
\usepackage{booktabs}
\usepackage{graphicx}
\usepackage{comment}
\usepackage{hyperref}
\usepackage{stfloats}
\usepackage{xcolor}

\newcommand{\ie}{\emph{i.e.}\xspace}
\newcommand{\eg}{\emph{e.g.}\xspace}
\newcommand{\etal}{\emph{et al.}\xspace}
\newcommand{\etc}{\emph{etc.}\xspace}
\newcommand{\vect}[1]{\mathbf{#1}}

\title{Is Joint Training Better for Deep Auto-Encoders?}

\author{Yingbo Zhou,
Devansh Arpit,
Ifeoma Nwogu,
Venu Govindaraju}

\begin{document}
\maketitle

\begin{abstract}
Traditionally, when generative models of data are developed via deep architectures, greedy layer-wise pre-training is employed. In a well-trained model, the lower layer of the architecture models the data distribution conditional upon the hidden variables, while the higher layers model the hidden distribution prior. But due to the greedy scheme of the layerwise training technique, the parameters of lower layers are fixed when training higher layers. This makes it extremely challenging for the model to learn the hidden distribution prior, which in turn leads to a suboptimal model for the data distribution. We therefore investigate joint training of deep autoencoders, where the architecture is viewed as one stack of two or more single-layer autoencoders. A single global reconstruction objective is jointly optimized, such that the objective for the single autoencoders at each layer acts as a local, layer-level regularizer. We empirically evaluate the performance of this joint training scheme and observe that it not only learns a better data model, but also learns better higher layer representations, which highlights its potential for unsupervised feature learning. In addition, we find that the usage of regularizations in the joint training scheme is crucial in achieving good performance. In the supervised setting, joint training also shows superior performance when training deeper models.  The joint training framework can thus provide a platform for investigating more efficient usage of different types of regularizers, especially in light of the growing volumes of available unlabeled data.
\end{abstract} 


\section {Introduction}
\IEEEPARstart{D}{eep} learning algorithms have been the object of much attention in the machine learning and applied statistics literature in recent years, due to the \textit{state-of-the-art} results achieved in various applications such as object recognition \cite{alex_imagenet,visual_deconvnet}, speech recognition \cite{dbn_speech,speech_mcrbm,speech_ae}, face recognition \cite{deepface}, etc. With the exception of convolutional neural networks (CNN), the training of multi-layered networks was in general unsuccessful until 2006, when breakthroughs were made by three seminal works in the field -  Hinton \etal \cite{DBLP:journals/neco/HintonOT06}, Bengio \etal \cite{DBLP:conf/nips/BengioLPL06} and Ranzato \etal \cite{Poultney06efficientlearning}.
They addressed the notion of \emph{greedy layer-wise pre-training} to initialize the weights of an entire network in an unsupervised manner, followed by a supervised back-propagation step. The inclusion of the unsupervised pre-training step appeared to be the missing ingredient which then lead to significant improvements over the conventional training schemes.

The recent progress in deep learning is more towards \emph{supervised} learning algorithms \cite{alex_imagenet,dropout12, dropconnect}.
While these algorithms obviate the need for the additional pre-training step in supervised settings, large amount of labeled data is still critical. Given the ever-growing volumes of unlabeled data and the cost of labeling, it remains a challenge to develop better unsupervised learning techniques to exploit the copious amounts of unlabeled data.

The unsupervised training of a network can be interpreted as learning the data distribution $p(\mathbf{x})$ in a probabilistic generative model of the data. A typical method for accomplishing this is to decompose the generative model into a latent conditional generative model and a prior distribution over the hidden variables.  When this interpretation is extended to a deep neural network \cite{DBLP:journals/neco/HintonOT06, dbn_rbm_theory, layerwise_theory}, it implies that while the lower layers model the conditional distribution $p(\mathbf{x}|\mathbf{h})$, the higher layers model the distribution over the latent variables. It has been shown that local-level learning accomplished via pre-training is important when training deep architectures (initializing every layer of an unsupervised deep Boltzmann machine (DBM) improves performance \cite{SalakhutdinovH09dbm}; similarly, initializing every layer of a supervised multi-layer network also improves performance \cite{DBLP:conf/nips/BengioLPL06}). But this greedy layer-wise scheme has a major disadvantage -- the higher layers have significantly less knowledge about the original data distribution than the bottom layers.  Hence, if the bottom layer does not capture the data representation sufficiently well, the higher layers may learn something that is not useful.  Furthermore, this error/bias will propagate through layers, \ie the higher the layer, the more the errors it will incur.  To summarize, the greedy layer-wise training focuses on the local constraints introduced by the learning algorithm (such as in auto-encoders), but loses sight of the original data distribution when training higher layers. To compensate for this disadvantage, all the levels of the deep architecture should be trained simultaneously. But this type of joint training can be very challenging and if done naively, will fail to learn \cite{SalakhutdinovH09dbm}.

Thus in this paper, we present an effective method for jointly training a multi-layer auto-encoder from end-to-end in an unsupervised fashion, and analyze its performance against greedy layer-wise training in various settings. This unsupervised joint training method consists of a global reconstruction objective, thereby learning a good data representation without losing sight of the original input data distribution.  In addition, it can also easily cope with local regularization on the parameters of hidden layer in a similar way as in layer-wise training, and therefore allow us to use more powerful regularizations proposed more recently.  This method can also be viewed as a generalization of single- to multi-layer auto-encoders.  Because our approach achieves a global reconstruction in a deep network, the feature representations learned are better. This attribute also makes it a good feature extractor and we confirm this by our extensive analysis. The representations learned from joint training approach consistently outperform those obtained from greedy layer-wise pre-training algorithms in unsupervised settings.  In supervsied setting, this joint training scheme also demonstrate superior performance for deeper models.



\subsection{Motivation}
Because supervised methods typically require large amounts of labeled data and the cost of acquiring these labels can be an expensive and time consuming task, it remains a challenge to continue to develop improved unsupervised learning techniques that can exploit large volumes of unlabeled data. Although the greedy layerwise pre-training procedure has till date been very successful, from an engineering perspective, it can be challenging to train, and monitoring the training process can be difficult. For the layerwise method, apart from the bottom layer where the unsupervised model is learning directly on the input, the training cost are measured with respect to the layer below. Hence, any changes in error values from one layer to the next has little meaning to the user. But by having one global objective, the joint training technique has training cost that are consistently measured with respect to the input layer. This way, one can readily monitor the changes in training errors and even with respect to post-training tasks such as classification or prediction.

Also, as stated earlier, the unsupervised training of a network can be interpreted as learning the data distribution $p(\mathbf{x})$ in a probabilistic generative model of the data. But in order for the unsupervised method to learn $p(\mathbf{x})$, a common strategy is to decompose $p(\mathbf{x})$ into $p(\mathbf{x}|\mathbf{h})$ and $p(\mathbf{h})$ \cite{layerwise_theory,dbn_rbm_theory}.  There are now two models to optimize, the conditional generating model $p(\mathbf{x}|\mathbf{h})$ and the prior model $p(\mathbf{h})$.  Since $p(\mathbf{h})$ covers a wide range of distributions and is hard to optimize in general, one tends to emphasize more on the optimization of $p(\mathbf{x}|\mathbf{h})$ and assume that the later learning of $p(\mathbf{h})$ could compensate for the loss occurred due to imperfect modeling.  Note that the prior $p(\mathbf{h})$ can also be decomposed in exactly the same way as $p(\mathbf{x})$, resulting in additional hidden layers.  Thus, one can recursively apply this trick and delay the learning of the prior.  The motivation behind this recursion is to expect that as the learning progresses through layers, the prior gets simpler, and thus makes the learning easier.  The greedy layer-wise training employs this idea, but it may fail to learn the optimum data distribution for the following reason:  in layerwise training, the parameters of the bottom layers are fixed after training and the prior model can only observe $\mathbf{x}$ through its fixed hidden representations.  Hence, if the learning of $p(\mathbf{x}|\mathbf{h})$ does not preserve all informatin regarding $x$ (which is very likely), then it is not possible to learn the prior $p(\mathbf{h})$ that leads to the optimum model for the data distribution.

For these reasons, therefore, we explore the possibility of jointly training deep autoencoders where the joint training scheme makes the adjustments of $p(\mathbf{h})$ and $p(\mathbf{x}|\mathbf{h})$ (and consequently $p(\mathbf{h}|\mathbf{x})$) possible with respect to each other, thus alleviating the burden for both models.  As a consequence, the prior model $p(\mathbf{h})$ can now observe the input $\mathbf{x}$ making it possible to fit the prior distribution better.  Hence, joint training \emph{makes global optimization more possible}.
\section {Background} \label{sec:ae}
In this section, we briefly review the concept of autoencoders and some of its variants, and expand on the notion of the 
\emph{deep autoencoder}, the primary focus of this paper.

\subsection {Basic Autoencoders (AE)}
A basic autoencoder is a one-hidden-layer neural network \cite{backprop, autoencoder}, and its objective is to reconstruct the input using its hidden activations so that the reconstruction error is as small as possible.  It takes the input and puts it through an encoding function to get the encoding of the input, and then it decodes the encodings through a decoding function to recover (an approximation of) the original input.  More formally, let $\mathbf{x} \in \mathbb{R}^d$ be the input,
\begin{align*}
	\vect{h} & = f_e(\vect{x}) = s_e(\vect{W_e}\vect{x}+\vect{b_e})\\
	\vect{x_r} & = f_d(\vect{x}) = s_d(\vect{W_d}\vect{h}+\vect{b_d})
\end{align*}
where $f_e: \mathbb{R}^d\mapsto\mathbb{R}^h$ and $f_d: \mathbb{R}^h\mapsto\mathbb{R}^d$ are encoding and decoding functions respectively, $\mathbf{W_e}$ and $\mathbf{W_d}$ are the weights of the encoding and decoding layers, and $\mathbf{b_e}$ and $\mathbf{b_d}$ are the biases for the two layers.  $s_e$ and $s_d$ are elementwise non-linear functions in general, and common choices are sigmoidal functions like $\tanh$ or logistic.  For training, we want to find a set of parameters $\vect{\Theta} = \{\vect{W_e}, \vect{W_d}, \vect{b_e}, \vect{b_d}\}$ that minimize the reconstruction error $\mathcal{J}_{AE}(\vect{\Theta})$:
\begin{equation}
	\mathcal{J}_{AE}(\vect{\Theta}) = \sum_{\vect{x}\in\mathcal{D}}\mathcal{L}(\vect{x}, \vect{x_r})
	\label{eq:ae}
\end{equation}
where $\mathcal{L}: \mathbb{R}^d \times \mathbb{R}^d \mapsto \mathbb{R}$ is a loss function that measures the error between the reconstructed input $\vect{x_r}$ and the actual input $\vect{x}$, and $\mathcal{D}$ denotes the training dataset.  We can also \emph{tie} the encoding and decoding weights by setting the weights $\vect{W_d} = \vect{W_e}^{\top}$.  Common choices of $\mathcal{L}$ includes sum-of-squared-errors for real valued inputs, cross-entropy for binary valued inputs \etc.  However, this model has a significant drawback in that if the number of hidden units $h$ is greater than the dimensionality of the input data $d$, the model can perform very well during training but fail at test time because it trivially copied the input layer to the hidden one and then copied it back. As a work-around, one can set $h < d$ to force the model to learn something meaningful, but the performance is still not very efficient.

\subsection {Denoising Autoencoder (DAE)}
Vicent \etal \cite{VincentLBM08} proposed a more efficient way to overcome the shortcomings of the basic autoencoder, namely the denoising autoencoder.  The idea is to corrupt the input before passing it to the network, but still require the model to reconstruct the uncorrupted input.  In this way, the model is forced to learn representations that are useful since trivially copying the input will not optimize this denoising objective (equation \ref{eq:denoise}).  Formally, let $\vect{x_c} \sim q(\vect{x_c}|\vect{x})$ be the corrupted version of the input, where $q(\cdot|\vect{x})$ is some corruption process over the input $\vect{x}$, then the objective it tries to optimize is:
\begin{equation}
	\mathcal{J}_{DAE}(\vect{\Theta}) = \sum_{\vect{x}\in \mathcal{D}}\mathbb{E}_{\vect{x_c} \sim q(\vect{x_c},\vect{x})}[\mathcal{L}(\vect{x}, f_d \circ f_e(\vect{x_c}))]
	\label{eq:denoise}
\end{equation}

\begin{figure*}[tb!p!]
\centering
{
\renewcommand{\arraystretch}{0.3}
\begin{tabular}{|c|c|}
\hline
{\tiny}& {\tiny} \\
\includegraphics[width=0.5\textwidth]{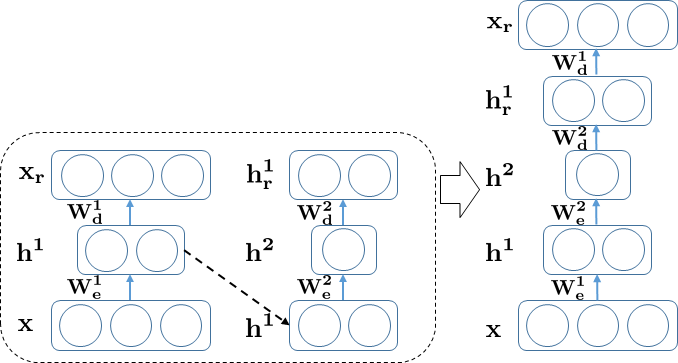} & \includegraphics[width=0.23\textwidth]{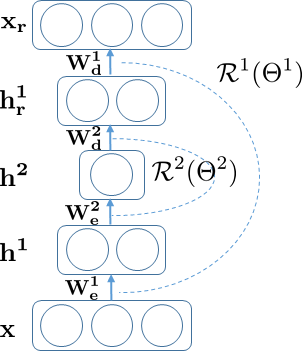}\\
\hline
\end{tabular}
}
\caption{Illustration of deep autoencoder that considered in this paper.  \textbf{Left:} an illustration of how greedy layerwise training is employed when training a two layered deep autoencoder: the bottom layer is first trained followed by a second layer using the previous layer's encoding as the input.  The loss of layerwise training is always measured with respect to its input, so the intermediate layers are trying to reconstruct back the representation (\ie $\vect{h^1}$) not the input data $\vect{x}$. The final deep autoencoder can be viewed as a stacking of encoding layers followed by decoding layers.  \textbf{Right:} the joint training scheme, where one start with a deep autoencoder and try to optimize it jointly to minimize the loss between the input and its final reconstruction.  Therefore, all the parameters are tuned towards better represent the input data.  In addition, regularization is enforced in each local layer to help learning.  Note that for simplicity all biases are excluded from the illustration.}
\label{fig:deep_ae}
\vspace{-10pt}
\end{figure*}

\subsection {Contractive Autoencoders (CAE)}
More recently, Rifai \etal \cite{RifaiVMGB11cae} proposed to add a special penalty term to the original autoencoder objective to achieve robustness to small local perturbations.  The penalty term was the Frobenius norm of the Jacobian matrix of the hidden activations with respect to the input, and their modified objective is:
\begin{equation}
	\mathcal{J}_{CAE}(\vect{\Theta}) =
	\sum_{\vect{x}\in\mathcal{D}}
	\left\{\mathcal{L}(\vect{x}, \vect{x_r}) +
	\lambda \|\nabla_{\vect{x}}f_e(\vect{x})\|^2_F\right\}
    \label{eq:cae}
\end{equation}
This penalty term measured the change within the hidden activations with respect to the input. Thus, the penalty term alone would prefer hidden activations to stay constant when the input varies (\ie the Frobenius norm would be zero).  The loss term will be small only if the reconstruction is close to the original, which is not possible if the hidden activation does not change according to the input.  Therefore, the hidden representation that tries to represent the data manifold would be preferred, since otherwise we would have high costs in both terms.


%
\subsection{Deep Autoencoders} \label{sec:deepae}
A deep autoencoder is a multi-layer neural network that tries to reconstruct its input (see Figure \ref{fig:deep_ae}).  In general, an $N$-layer deep autoencoder with parameters $\vect{\Theta} = \{\vect{\Theta^i} | i \in \{1,2,\ldots N\}\}$, where $\vect{\Theta^i} = \{\vect{W_e^i}, \vect{W_d^i}, \vect{b_e^i}, \vect{b_d^i}\}$ can be formulated as follows:
\begin{align}
	& \vect{h^i} = f_e^i(\vect{h^{i-1}}) =
	 s_e^i(\vect{W_e^i}\vect{h^{i-1}}+\vect{b_e^i})\\
	& \vect{h_r^i} = f_d^i(\vect{h_r^{i+1}}) =
	 s_d^i(\vect{W_d^i}\vect{h_r^{i+1}}+\vect{b_d^i})\\
	& \vect{h^0} = \vect{x}
\end{align}
The deep autoencoder architecture therefore contains multiple encoding and decoding stages made up of a sequence of encoding layers followed by a stack of decoding layers. Notice that the deep autoencoder therefore has a total of $2N$ layers.  This type of deep autoencoder has been investigated in several previous works (see \cite{Hinton06science, Martens10HF, SutskeverMDH13momentum} for examples). A deep autoencoder can also be viewed as an unwrapped stack of multiple autoencoders with the higher layer autoencoder taking the lower layer encoding as its input.  Hence, when viewed as a stack of autoencoders, one could train the stack from bottom to top.

\subsection{General Autoencoder Objective}
By observing the formulations of Equations (\ref{eq:ae}-\ref{eq:cae}), we can rewrite the single layer autoencoder objective in a more general form as: 
\begin{equation}
\mathcal{J}_{GAE} (\vect{\Theta}) =
\sum_{\vect{x}\in\mathcal{D}}\mathbb{E}_{\vect{x_c} \sim Q(\vect{x_c},\vect{x})}[\mathcal{L}(\vect{x}, f_d \circ f_e(\vect{x_c}))] +
\lambda \mathcal{R}(\vect{\Theta})
\label{eq:gae}
\end{equation}
where $\lambda \in [0, +\infty)$, $Q(\vect{x_c}|\vect{x})$ is some conditional distribution over the input and $Q(\vect{x_c},\vect{x}) = Q(\vect{x_c}|\vect{x})P(\vect{x})$, and $\mathcal{R}(\vect{\Theta})$ is an arbitrary regularization function over the parameters.  The choice of a good regularization term $\mathcal{R}(\cdot)$ seems to be the ingredient that has led to the recent successes of several autoencoder variants (the models proposed in paper \cite{XieXC12,ChenWSB14Marginal-dae}, for example).

It is straightforward to see that we can recover the previous autoencoder objectives from Equation \ref{eq:gae}. For example, the basic autoencoder can be obtained by setting $Q$ to be a Dirac delta at the data point, and $\lambda=0$.  Since this objective is more general, going forward, we will refer to this objective rather than the specific autoencoder ones presented earlier.

\section {Our Joint Training Method}
\label{sec:method}
As mentioned previously, a deep autoencoder can be optimized via layer-wise training, \ie bottom layer objective $\mathcal{J}_{GAE}(\vect{\Theta^i})$ is optimized before $\mathcal{J}_{GAE}(\vect{\Theta^{i+1}})$ (\ie from bottom to top, see also Figure \ref{fig:deep_ae} left) until the desired depth is reached.  If one would like to jointly train all layers, a natural question would be -- \emph{why not combine these objectives into one and train all the layers jointly?}  Directly combing all the objectives (\ie optimizing $\sum_i \mathcal{J}_{GAE}(\vect{\Theta^i})$) is not appropriate, since the end goal is to reconstruct the input as accurate as possible, and not the intermediate representations.  The aggregated loss and regularization may hinder the training process since the objective has deviated from the goal. Furthermore, during training, the representation of the lower layers is varying continuously (since the parameters are changing), making the optimization very difficult\footnote{We have tried this scheme on several datasets but the results are very poor.}.

So, focusing on the goal of reconstructing the input, we explore the following {\bf joint training objective} for an $N$-layered deep autoencoder:
\begin{equation}
\mathcal{J}_{Joint}(\vect{\Lambda}) =
\sum_{\vect{x}\in\mathcal{D}}\mathbb{E}_{Q(\vect{x}, \vect{h_c^0},\ldots,\vect{h_c^N})}[\mathcal{L}(\vect{x}, \vect{x_r})] +
\sum_{i=1}^N \lambda^i \mathcal{R}^i(\vect{\Theta^i})
\label{eqn:joint_obj}
\end{equation}

where $\vect{\Lambda}=\{\cup_{i=1}^N\vect{\Theta^i}\}$, $Q(\vect{x}, \vect{h_c^0},\ldots,\vect{h_c^N}) = P(\vect{x})\prod_{i=1}^NQ^i(\vect{h_c^i}|\vect{h^i})P(\vect{h^i})$; and $\vect{h^i_c} \sim Q^i(\vect{h^i_c}|\vect{h^i})$
with $\vect{h_0} = \vect{x}$, and $\vect{x_r} = f_d^1 \circ f_d^2 \circ \cdots \circ f_d^N(h^N)$.

In other words, this method is optimizing the reconstruction error directly from a multi-layer neural network, and at the same time enabling us to conveniently apply more powerful regularizers for the single autoencoders at each layer.  For example, if we want to train a two-layered denoising autoencoder using this method, we will need to corrupt the input and feed it into the first layer, and then corrupt the hidden output from the first layer and feed it into the second layer; then we reconstruct from the second layer hidden activations followed by a reconstruction to the original input space using the first layer reconstruction weight; we then measure the loss between the reconstruction and the input and do gradient update to all the parameters (see Figure \ref{fig:deep_ae}).

In this formulation, we not only train all layers jointly by optimizing one global objective (the first part of equation \ref{eqn:joint_obj}), so that all hidden activations will adjust according to the original input; but we also take into account the local constraints of the single layer autoencoders (the second part of equation \ref{eqn:joint_obj}).  Locally, if we isolate each autoencoder out from the stacks and look at them individually, this training process is optimizing almost exactly the single layer objective as in layerwise case, but without the unnecessary constraint on reconstructing back the intermediate representations.  Globally, because the loss is measured between the reconstruction and the input, all parameters must be tuned to minimize the global loss, and thus the resulting higher level hidden representations would be more representative of the input.



This approach addresses some of the drawbacks of layerwise training: since all the parameters are tuned simultaneously toward the reconstruction error of the original input, the optimization is no longer local for each layer.  In addition, the reconstruction error in the input space makes the training process much easier to monitor and interpret.  To sum up, this formulation provides an easier and more modular way to train deep autoencoders.

We now relate this joint objective to some of the existing, well-known deep learning paradigms that implement deterministic objective functions. We also relate this work to the techniques in the literature that explain deep learning from a probabilistic perspective.

\subsection{Relationship to Weight Decay in Multi-layer Perceptron}
It is easy to see that if we replace all the regularizers $\mathcal{R}^i(\vect{\Theta}^i)$ with $L_1$ or $L_2$ norms, and set $Q^i(\tilde{\vect{z}} | \vect{z})$ to Dirac delta distribution at point $\vect{z}$ in Equation (\ref{eqn:joint_obj}),
%
%
we recover the standard multi-layer perceptron (MLP) with $L_1$ or $L_2$ weight decay with one exception -- MLP is not commonly used for unsupervised learning.  So if we replace the loss term with some supervised loss, it is then identical to the ordinary MLP with corresponding weight decay.

\subsection{Relationship to Greedy Layerwise Pre-training}
It is straight forward to see that in the single layer case, the proposed training objective in Equation (\ref{eqn:joint_obj}) is equivalent to greedy layerwise training.  It would be interesting to investigate the relationship of this method and layerwise training in multi-layer cases.  Therefore, we construct a scenario that make these two method very similar by modify the joint training algorithm slightly by setting a training schedule with learning rates $\alpha^i_t$, and regularizer $\lambda^i_t$, where $i,t \in \mathbb{Z}^+$, $i$ and $t$ indicate the corresponding layer and training iteration.  The objective in equation \ref{eqn:joint_obj} and the gradient update is as follows:
\begin{align}
\mathcal{J}_{S}(\vect{\Lambda}_t) &=
\sum_{\vect{x}\in\mathcal{D}}\mathbb{E}_{Q(\vect{x}, \vect{h_c^0},\ldots,\vect{h_c^N})}[\mathcal{L}(\vect{x}, \vect{x_r})] +
\sum_{i=1}^N \lambda^i_t \mathcal{R}^i(\vect{\Theta}^i)\\
\vect{\Theta}^i_{t+1} &\leftarrow \vect{\Theta}^i_{t} - \alpha^i_t \Delta\vect{\Theta}^i_{t}
\end{align}
Let $\alpha_t^i \in \{0, \epsilon\}$ and $\sum_{i} \alpha_t^i = \epsilon$, where $\epsilon \in \mathbb{R}^+$ indicates the learning rate.  Let us set the value of $\alpha$ in the following way: $\alpha_t^1 = \epsilon$ for $t \in [1,T]$, $\alpha_t^2 = \epsilon$ for $t \in [T+1,2T]$, $\ldots, \alpha_t^N = \epsilon$ for $t \in [(N-1)T+1,NT]$, where $T$ is the number of iterations used in the greedy layerwise training.

In this way, the joint training scheme is set up in a very similar way to the layerwise training scheme, \ie only tune the parameter of a single layer at a time from bottom to top.  However, since the loss is measured in the domain of $\vect{x}$ instead of $\vect{h}_i$, the learning will still behave differently.  This will be more apparent as we write down both joint loss ($\mathcal{L}_J$) and layerwise loss ($\mathcal{L}_L$) for training a given layer $j$ as follows:
\begin{align}
\label{eqn:joint_loss}
& \mathcal{L}_J(\vect{x}, \vect{x_r}) = \mathcal{L}(f_d^{*1} \circ \cdots \circ f_d^{*j-1}\circ f_d^{j} \circ f_e^{j}(\vect{h}_{j-1}), \vect{x}) \\
\label{eqn:layerwise_loss}
& \mathcal{L}_L(\vect{h}_{j-1}, \vect{h_r}_{(j-1)}) = \mathcal{L}(f_d^{j} \circ f_e^{j}(\vect{h}_{j-1}), \vect{h}_{j-1})
\end{align}
where $f_d^{*i}$ and $f_e^{*i}$ represent the trained decoding and encoding function respectively.  In other words, we do not optimize the parameters of these function during training.  Note that, the loss and the resulting parameter update will be very different between equation \ref{eqn:joint_loss} and \ref{eqn:layerwise_loss} in general, since the gradient for the joint loss will be modified by the previous layers parameters but not in the layerwise case.  It is somewhat surprising that even by constructing a very similar situation the two methods still behave very differently.  Note that even in the special case where one uses linear activation (which is not very practical), these two losses are still not equivalent.  Hence, the joint training will perform very differently from layerwise in general.

%
%
\subsection{Relationship to Learning the Data Distribution}
If we take a probabilistic view of the learning of a deep model as learning the data distribution $p(\vect{x})$, a common interpretation of the model is to decompose $p(\vect{x})$ into:
\begin{equation}
p(\vect{x}) = \sum_{\vect{h}}p(\vect{x}|\vect{h})p(\vect{h}) = \sum_{\vect{h}}p(\vect{x}|\vect{h})\sum_{\vect{y}}p(\vect{h}|\vect{y})q(\vect{y})
\end{equation}
where $q(\vect{y})$ is the empirical distribution over data.  So the model is decomposed such that the bottom layer models the distribution $p(\vect{x}|\vect{h})$ and the higher layers models the prior $p(\vect{h})$.  Notice that, if we apply layerwise training it is only possible to learn the prior $p(\vect{h})$ through a fixed $p(\vect{h}|\vect{x})$, and thus the prior will not be optimal with respect to $p(\vect{x})$ if $p(\vect{h}|\vect{x})$ does not preserve all information regarding $\vect{x}$.  On the other hand, in joint training, both the generative distribution $p(\vect{x}|\vect{h})$ and the prior $p(\vect{h})$ are tuned together, and therefore is more likely to obtain better estimations of the true $p(\vect{x})$.  In addition, while training layers greedily, we are not taking into account the fact that some more capacity may be added later to improve the prior for hidden units.  This problem is also alleviated by joint training since the architecture is fixed at the beginning of training and all the parameters are tuned towards better representation of the data.

\subsection{Relationship to Generative Stochastic Networks}
Bengio \etal \cite{GSN} recently proposed a new alternative to maximum likelihood for training probabilistic models -- the generative stochastic networks (GSNs).  The idea is that learning the data distribution $p(\vect{x})$ directly is hard since it is highly multi-modal in general. On the other hand, one can try to learn to approximate the Markov chain transition operator (\eg $p(\vect{h}|\vect{h_t}, \vect{x_t})$, $p(\vect{x}|\vect{h_t})$).  The intuition is that the move in Markov chain are mostly local, and thus these distributions are likely to be less complex or even unimodal, make it an easier learning problem.  For example, the denoising autoencoder learns $p(\vect{x}|\vect{\tilde{x}})$ where $\vect{\tilde{x}} \sim c(\vect{\tilde{x}}|\vect{x})$ is a corrupted example.  They show that if one can get a consistent estimator of $p(\vect{x}|\vect{h})$, then following the implied Markov chain, the stationary distribution of this chain will converge to the true data distribution $p(\vect{x})$.

Like the denoising autoencoder, a deep denoising autoencoder also defines the Markov chain:
\begin{align}
& \tilde{X}_{t+1} \sim c(\tilde{X}|X_t) \\
& X_{t+1} \sim p(X|\tilde{X}_{t+1})
\end{align}
Therefore, a deep denoising autoencoder also learns a data generating distribution within the GSN framework, and we can generate samples from it. 
\section {Experiments and Results}
\label{sec:experiment}
In this section, we empirically analyse the unsupervised joint training method with the following questions: 1) does joint training lead to better data models? 2) does joint training result in better representations that would be helpful for other tasks? 3) what role does the more recent usage of regularizers for autoencoder play in joint training? 4) does joint training affect the performance of supervised finetuning?

We tested our approach on MNIST \cite{mnistlecun} -- a digit classification dataset contains 60,000 training and 10,000 test examples, where we used the standard 50,000 and 10,000 training and validation split.  In addition, we also used the MNIST variation datasets \cite{LarochelleECBB07} each with 10,000 data points for training, 2,000 for validation and 50,000 for test.  Additional shape datsets employed in \cite{LarochelleECBB07} are also employed, this set of datasets contains two shape classification tasks.  One is to classify short and tall rectangles, the other is to classify convex and non-convex shapes.  All of these datasets have 50,000 testing examples.  The rectangle dataset has 1,000 and 200 training and validation examples respectively, and the convex dataset has 6,000 and 2,000 training and validation respectively.  The rectangle dataset also has a variation that uses image as the foreground rectangles, and it has 10,000 and 2,000 training and validation examples, respectively (see Figure \ref{fig:samples} left for visual examples from these datasets).

We tied the weights of the deep autoencoders (\ie $\vect{W_e}^i = \vect{W_d}^{i\top}$) in this set of experiments, and set each layer with 1,000 hidden units using logistic activations, and cross-entropy loss was applied as the training cost.  We optimized the deep network using rms-prop \cite{rms-prop} with 0.9 decay factor for the rms estimate and 100 samples per mini-batch.  The hyper-parameters were chosen on the validation set, and the model that obtained best validation result was used to obtain the test set result.  The hyper-parameters we considered were the learning rate (from the set \{0.001,0.005,0.01,0.02\}), noise level (from the set \{0.1,0.3,0.5,0.7,0.9\}) for deep denoising autoencoders (deep-DAE), and contraction level (from the set \{0.01,0.05,0.15,0.3,0.6\}) for deep contractive autoencoders (deep-CAE).  Gaussian noise is applied for DAE.
\subsection{Does Joint Training Lead to Better Data Models?}
As mentioned in previous sections, we hypothesize that \emph{the joint training will alleviate the burden on both the bottom layer distribution $p(x|h)$ and top layer priors $p(h)$, and hence result a better data model $p(x)$}.  In this experiment, we inspect the goodness on the modeling of data distribution through samples from the model.  Since the deep denoising autoencoder follows the GSN framework, we can follow the implied Markov chain to generate samples.  The models are trained for 300 epochs using both layerwise and joint training method\footnote{Each layer is trained for 300 epochs in layerwise training}, and the quality of samples are then estimated by measuring the log-likelihood of the test set under a Parzen window density estimator \cite{DBLP:journals/corr/abs-1012-3476, Breuleux+al-TR-2010}.  This measuerment can be seen as a lower bound on the true log-likelihood, and will converge to the true likelihood as the number of samples increase and with an appropriate Parzen window parameter. 10,000 consecutive samples were generated for each of the datasets with models that were trained using the layerwise and joint training method, and we used a Gaussian Parzen window for the density estimation.  The estimated log-likelihoods on the respective test sets are shown in Table \ref{tbl:likelihood}\footnote{We note that this estimate has a little high variance, but this is to our knowledge the best available method for estimating generative models that can generate samples but not estimate data likelihood directly.}.  We use a suffix 'L' and 'J' to denote the layerwise and joint training method, respectively.

\begin{table*}[tbp]
\caption{Test set log-likelihood estimated from Parzen window density estimator.  Suffix 'L' is used to denote the model trained from layerwise method and 'J' denotes the model trained from the joint training method.}
\label{tbl:likelihood}
\centering
\begin{tabular}{l|rr|rr}
\toprule
Dataset/Method & \multicolumn{1}{c}{DAE-2L}                 & \multicolumn{1}{c|}{DAE-2J} & \multicolumn{1}{c}{DAE-3L}       & \multicolumn{1}{c}{DAE-3J}                \\
\midrule
MNIST          & 204$\pm$1.59          & \textbf{266$\pm$1.32}       & 181$\pm$1.68          & \textbf{270$\pm$1.26}  \\
basic          & 201$\pm$1.69          & \textbf{205$\pm$1.59}       & 183$\pm$1.62          & \textbf{217$\pm$1.51}  \\
rot            & \textbf{174$\pm$1.41} & \textbf{172$\pm$1.49}       & 163$\pm$1.60          & \textbf{187$\pm$1.36}  \\
bg-img         & \textbf{156$\pm$1.41} & \textbf{154$\pm$1.45}       & 142$\pm$1.56          & \textbf{155$\pm$1.48}  \\
bg-rand        & -267$\pm$0.38         & \textbf{-252$\pm$0.36}      & -275$\pm$0.37         & \textbf{-249$\pm$0.36} \\
bg-img-rot     & \textbf{151$\pm$1.46} & \textbf{152$\pm$1.50}       & 139$\pm$1.52          & \textbf{149$\pm$1.31}  \\
rect           & \textbf{160$\pm$1.08} & \textbf{161$\pm$1.12}       & \textbf{152$\pm$1.15} & \textbf{154$\pm$1.12}  \\
rect-img       & \textbf{275$\pm$1.35} & \textbf{273$\pm$1.30}       & 269$\pm$1.40          & \textbf{272$\pm$1.37}  \\
convex         & -967$\pm$8.99         & \textbf{-853$\pm$8.85}      & -1011$\pm$10.79       & \textbf{-704$\pm$8.50} \\
\bottomrule
\end{tabular}
\end{table*}

\begin{figure*}[htbp]
\centering
\begin{tabular}{ccc}
Training Samples & Layerwise Samples & Joint Training Samples \\
\includegraphics[width=0.1\textwidth]{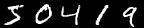} & \includegraphics[width=0.4\textwidth]{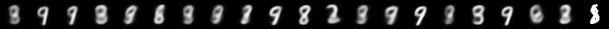} &  \includegraphics[width=0.4\textwidth]{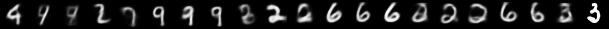} \\
\includegraphics[width=0.1\textwidth]{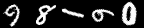} & \includegraphics[width=0.4\textwidth]{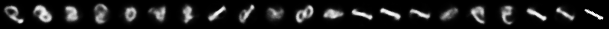} &  \includegraphics[width=0.4\textwidth]{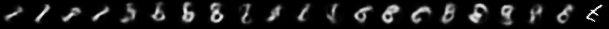} \\
\includegraphics[width=0.1\textwidth]{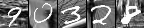} & \includegraphics[width=0.4\textwidth]{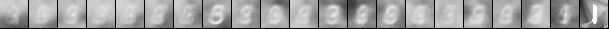} &  \includegraphics[width=0.4\textwidth]{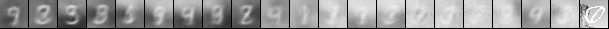} \\
\includegraphics[width=0.1\textwidth]{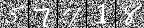} & \includegraphics[width=0.4\textwidth]{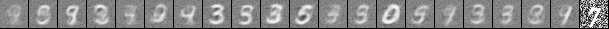} &  \includegraphics[width=0.4\textwidth]{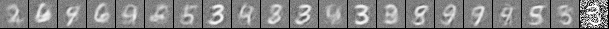} \\
\includegraphics[width=0.1\textwidth]{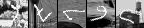} & \includegraphics[width=0.4\textwidth]{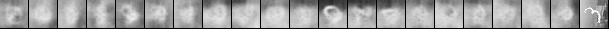} &  \includegraphics[width=0.4\textwidth]{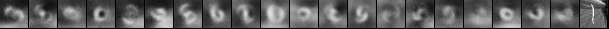} \\
\includegraphics[width=0.1\textwidth]{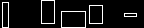} & \includegraphics[width=0.4\textwidth]{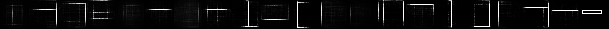} &  \includegraphics[width=0.4\textwidth]{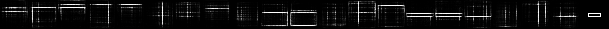} \\
\includegraphics[width=0.1\textwidth]{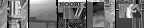} & \includegraphics[width=0.4\textwidth]{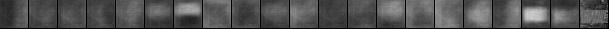} &  \includegraphics[width=0.4\textwidth]{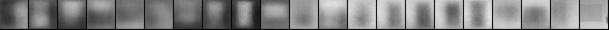} \\
\includegraphics[width=0.1\textwidth]{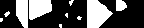} & \includegraphics[width=0.4\textwidth]{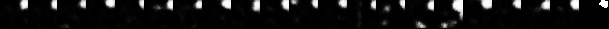} &  \includegraphics[width=0.4\textwidth]{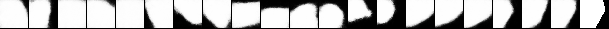} \\
\end{tabular}
\caption{Samples generated from deep denoising autoencoder with the best estimated log-likelihood trained on MNIST and the image classification datasets used in \cite{LarochelleECBB07}.  \textbf{From top to bottom:} samples from MNIST, MNIST-rotation, MNIST-bg-image, MNIST-bg-random, MNIST-bg-rot-image, rectangle, rectangle-image and convex dataset.  \textbf{Left:} training samples from the corresponding dataset.  \textbf{Middle:} consecutive samples generated from deep denoising autoencoder trained by using layerwise scheme.  \textbf{Right:} consecutive samples generated from deep denoising autoencoder trained by joint training.  Notice that only every fourth sample is shown, the last column of each consecutive samples shows the closest image from the training set to illustrate that the model is not memorizing the training data.  See Figure \ref{fig:expanded_samples} for samples from longer runs.}
\label{fig:samples}
\end{figure*}

For qualitative purposes the generated samples from each dataset are shown in Figure \ref{fig:samples}; the quality of samples are comparable in both cases, however, the models trained through joint training shows faster mixing with less spurious samples in general.  It is also interesting to note that, in most of the cases (see Table \ref{tbl:likelihood}) the log-likelihood on the test set improved with deeper models in the joint training case, whereas in layerwise settings, the likelihood dropped with additional layers.  This illustrates one advantage of using joint training scheme to model data since it accommodates the additional capacity of the hidden prior $p(h)$ while training the whole model.
\begin{figure*}[!tbph]
\centering
\begin{tabular}{ccc}
\includegraphics[width=0.3\textwidth]{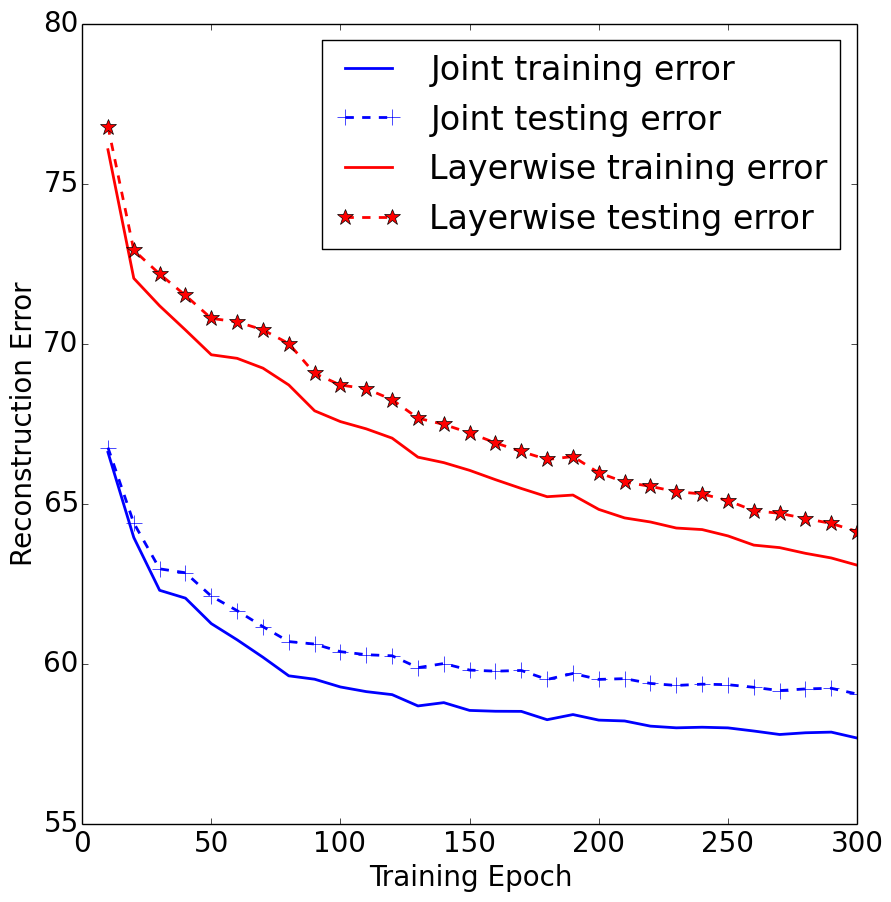} &  
\includegraphics[width=0.3\textwidth]{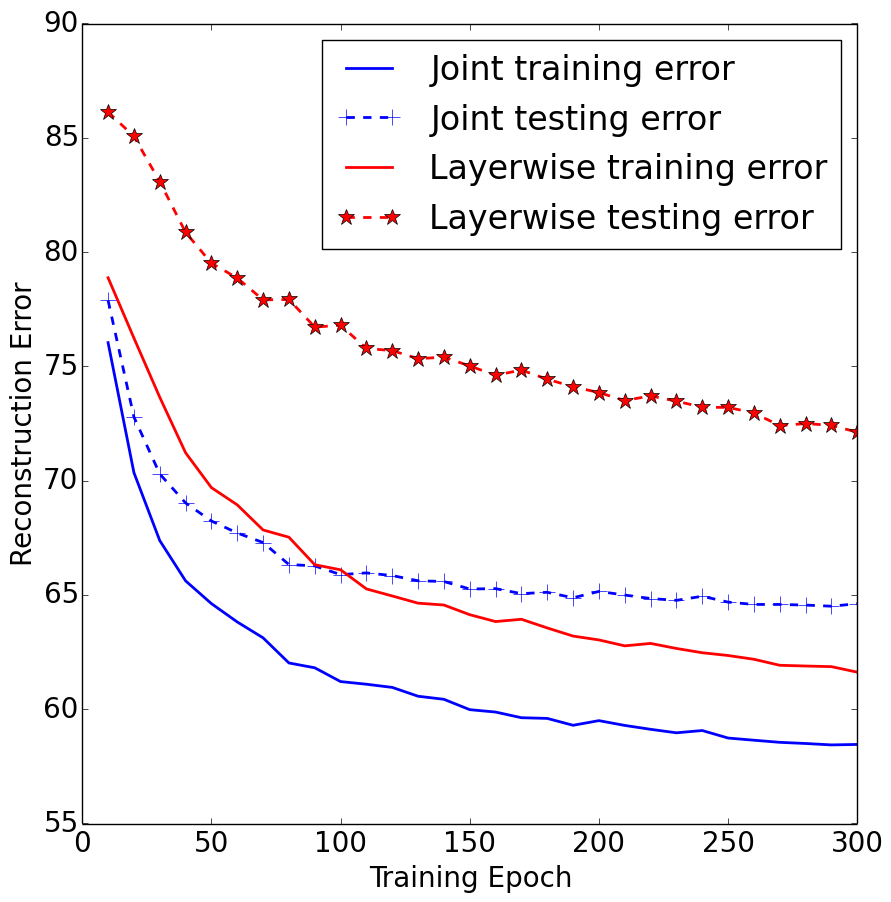} &
\includegraphics[width=0.3\textwidth]{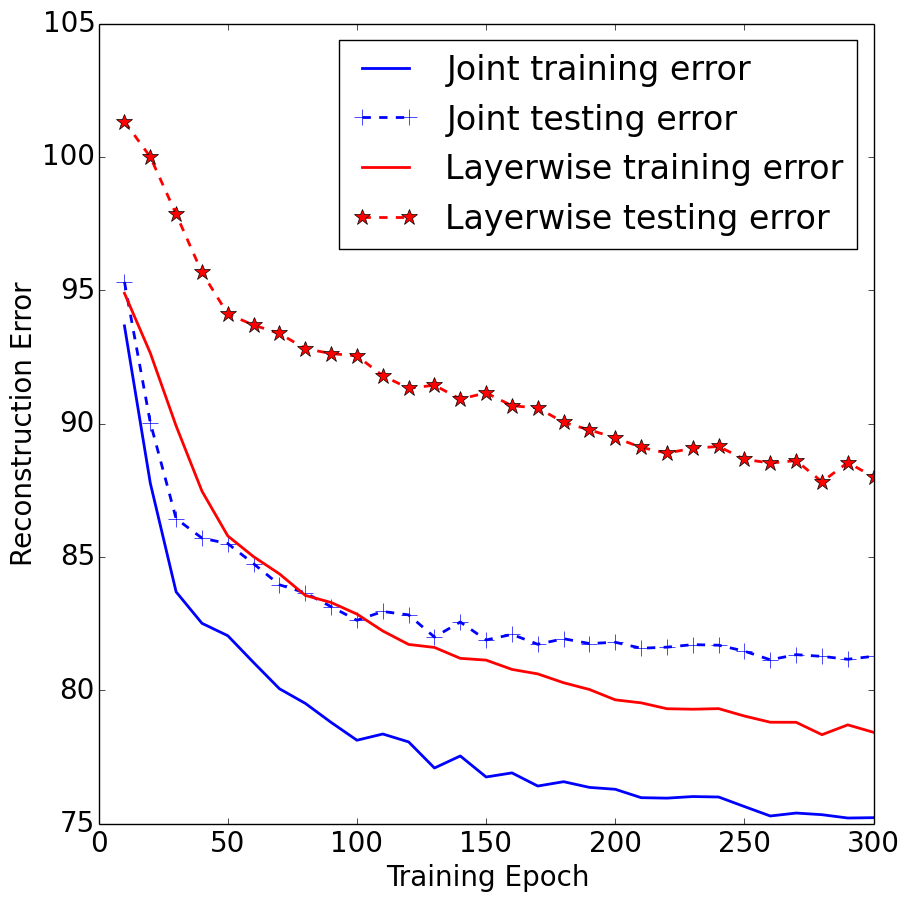} \\
(a) & (b) & (c) \\
\includegraphics[width=0.3\textwidth]{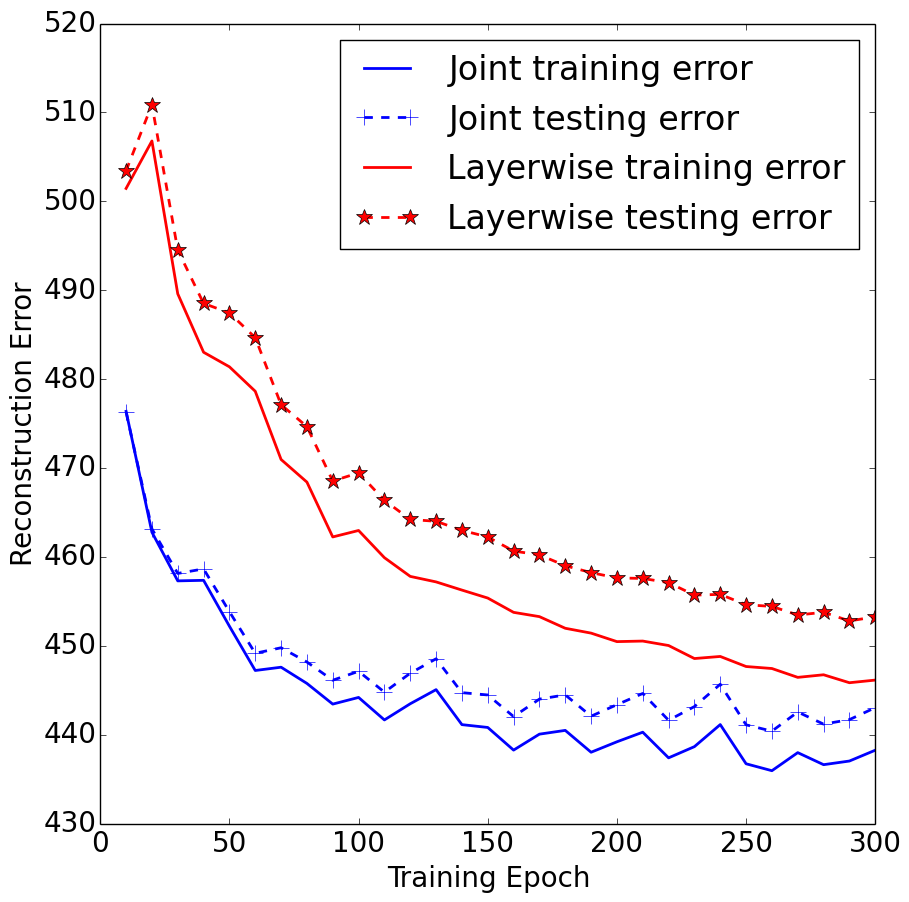} & 
\includegraphics[width=0.3\textwidth]{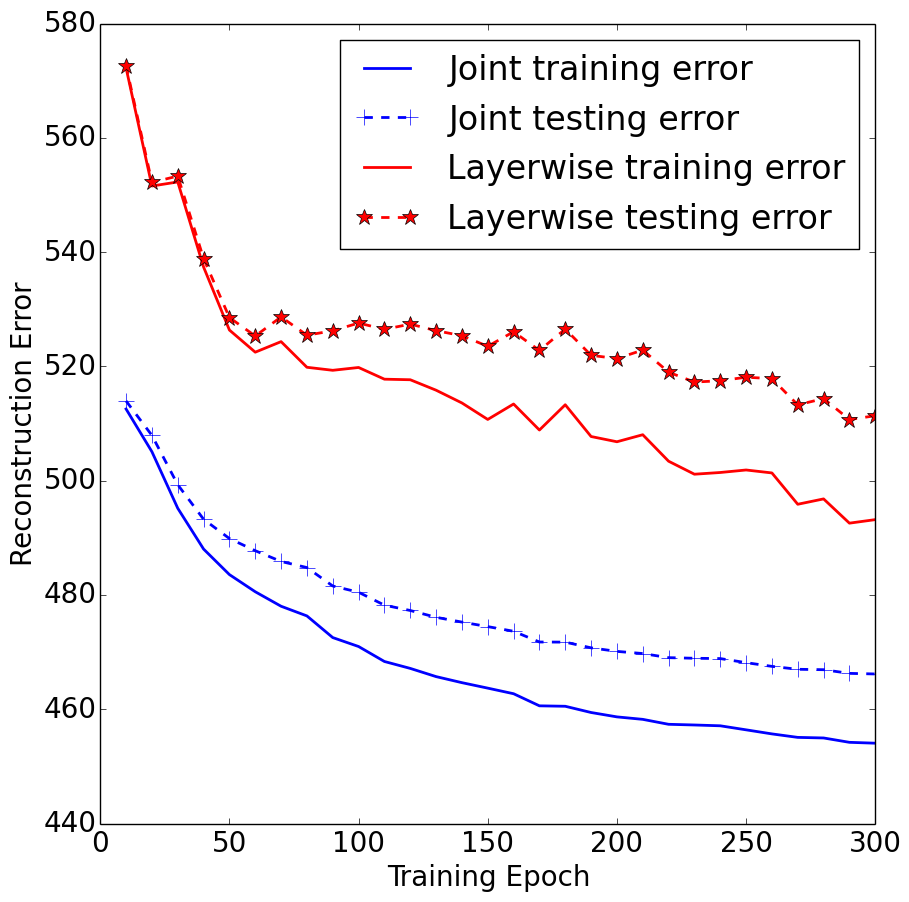} &
\includegraphics[width=0.3\textwidth]{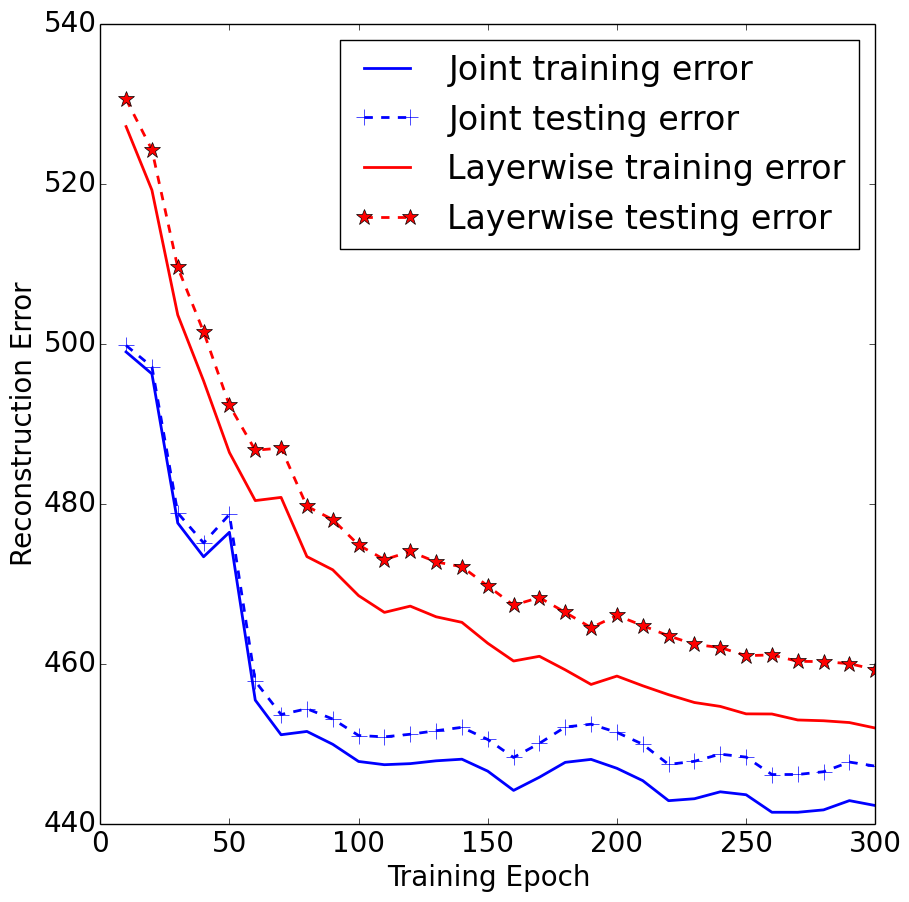} \\ 
(d) & (e) & (f) \\
\includegraphics[width=0.3\textwidth]{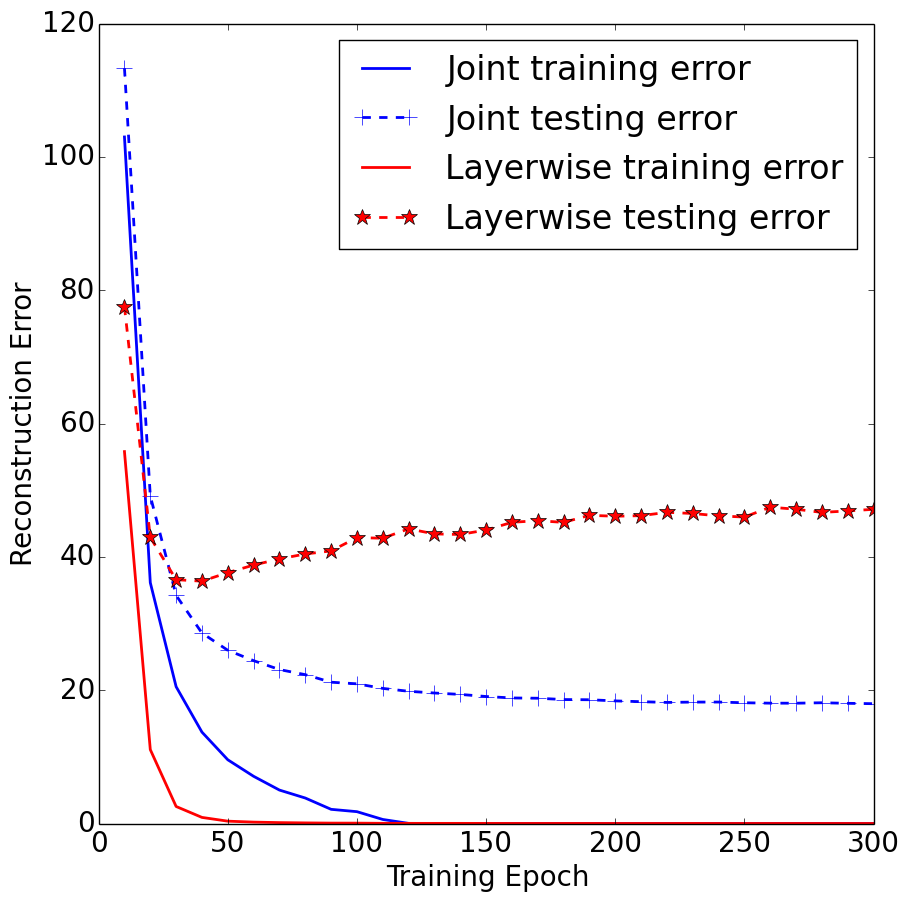} &
\includegraphics[width=0.3\textwidth]{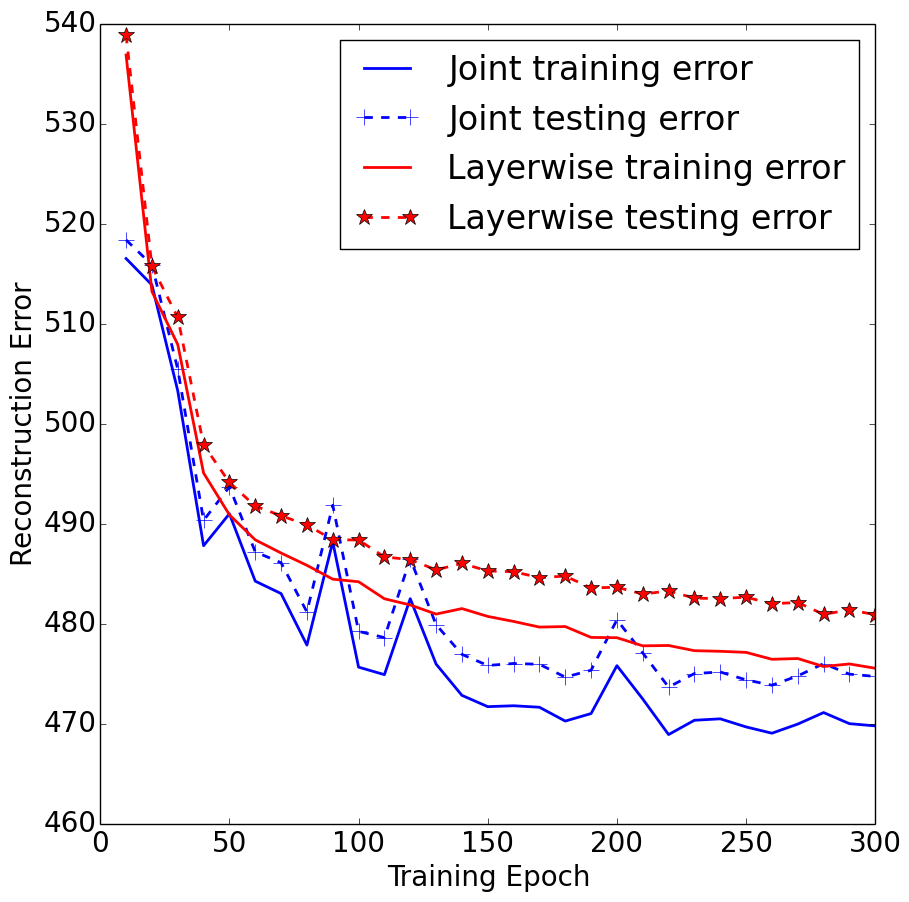} & 
\includegraphics[width=0.3\textwidth]{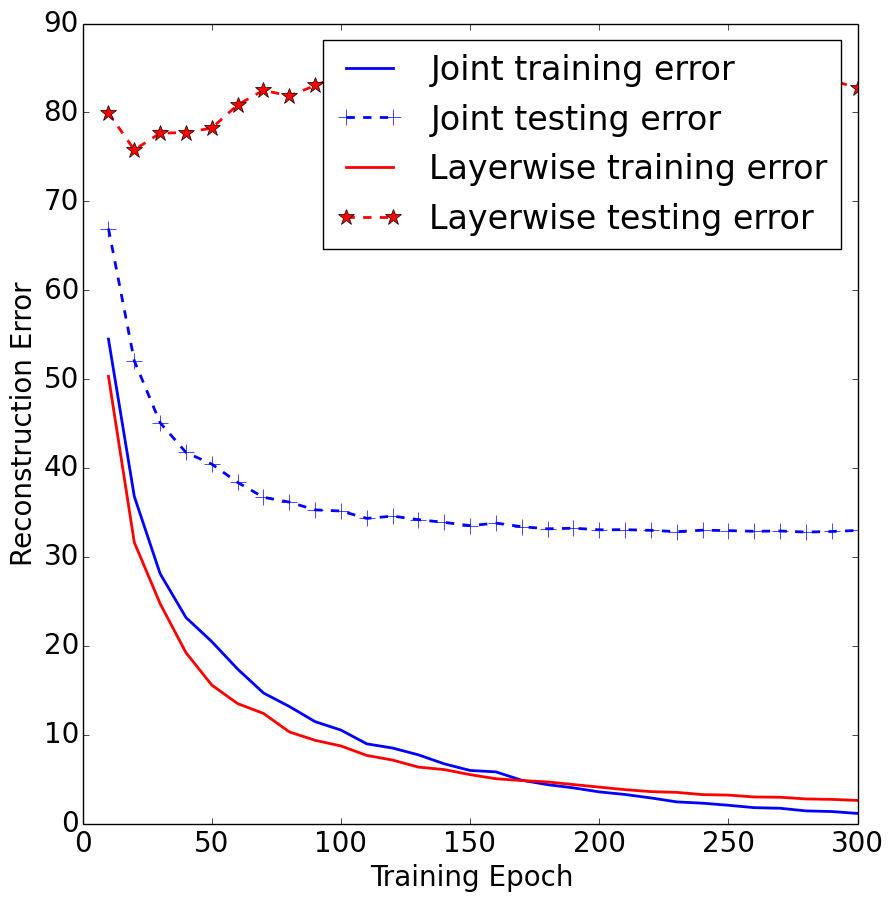} \\
(g) & (h) & (i) \\
\end{tabular}
\caption{Training and testing reconstruction errors from 2-layered deep denoising autoencoder on various datasets using layerwise and joint training.  a) MNIST, b) MNIST-basic, c) MNIST-rotation, d) MNIST-bg-image, e) MNIST-bg-random, f) MNIST-bg-rot-image, g) rectangle, h) rectangle-image and i) convex dataset.}
\label{fig:rec_errors}
\end{figure*}

Since the joint training objective in eq. \ref{eqn:joint_obj} focuses on reconstruction, it is expected that the reconstruction errors from models trained by joint training should be less than that obaited from layerwise training.  To confirm this we also record the training and testing errors as training progresses.  As can be seen from Figure \ref{fig:rec_errors}, it is clearly the case that joint training achieves better performance in all cases\footnote{We did not show the rest models since the trends are very similar.}.  It is also interesting to note that the models from joint training are less prone to overfitting as compare to layerwise case, this is true even in case of less training examples (see Figure \ref{fig:rec_errors}g and i). 

%
%
\subsection{Does Joint Training Result in Better Representations?}
In the previous experiment we illustrated the advantage of joint training over greedy layerwise training when learning the data distribution $p(\vect{x})$.  We also confirmed that joint training has better reconstruction error in both training and testing as compare to layerwise case, since it focuses on reconstruction.  A natural follow-up question would be to find out \emph{if tuning all the parameters of a deep unsupervised model towards better representing the input, lead to better higher level representations}. To answer this question the following experiment was conducted.  We trained two- and three-layered deep autoencoder\footnote{Note that the actual depth of the network has double the number of hidden layers because of the presence of intermediate reconstruction layers.} using both the layerwise and our joint training methods for 300 epochs\footnote{Each layer is trained for 300 epochs in layerwise training.}.  We then fix all the weights and used the deep autoencoders as feature extractors on the data, \ie use the top encoding layer representation as the feature for the corresponding data.  A support vector machine (SVM) with a linear kernel was further trained on those features and the test set classification performance was evaluated.  The model that achieved the best validation performance was used to report the test set error, and the performance for the two different models is shown in Tables \ref{tbl:dae_unsup} and \ref{tbl:cae_unsup}.  We report the test set classification error (all in percentages) along with its 95\% confidence interval for all classification experiments.  Since the computation of contractive penalty of higher layers with respect to the input is expensive for the deep contractive autoencoder, we set $\mathcal{R}^i(\Theta^i) = \|\nabla_{\vect{h}^{i-1}}\vect{h}^{i}\|^2_F$ to save some computations.  The result from Tables \ref{tbl:dae_unsup} and \ref{tbl:cae_unsup} suggest that the representations learned from the joint training was generally better.  It is also important to note that the model achieved low error rate without any finetuning by using the labeled data; in other words, the feature extraction was completely unsupervised.  Interestingly, the three-layered models seemed to perform worse than two-layered models in most of the cases for all methods, which contradicts the generative performance.  This is because the goal for a good unsupervised data model is to capture all information regarding $p(\vect{x})$, whereas for supervised tasks the goal is to learn a good model over $p(\vect{y}|\vect{x})$.  $p(\vect{x})$ contains useful information regarding $p(\vect{y}|\vect{x})$ but not all the information from $p(\vect{x})$ might be helpful.  Therefore, good generative performance does not necessarily translate to good discriminative performance.
\begin{table*}[!t!b!p!h]
\caption{Linear SVM classification performance on top-level representations learnt unsupervised by deep denoising autoencoders trained from various schemes. Suffix `L' is used to denote models that are trained from layerwise method, `U' is used to denote models that are jointly trained without regularization but initialized with pre-trained weights from layerwise scheme, `J' is used to denote models that are jointly trained with corresponding regularizations with randomly initialized parameters, and `UJ' is used to denote models that are jointly trained with corresponding regularizations and initialized with pre-trained weights from layerwise scheme. See text for details.}
\label{tbl:dae_unsup}
\centering
\begin{tabular}{l|cccc|cccc}
\toprule
Dataset/Method  &  DAE-2L  &  DAE-2U  &  DAE-2J  &  DAE-2UJ  &  DAE-3L  &  DAE-3U  &  DAE-3J  &  DAE-3UJ  \\
\midrule
MNIST  &  \textbf{1.48 $\pm$ 0.24} &  1.94 $\pm$ 0.27 &   \textbf{1.39 $\pm$ 0.23} &   \textbf{1.47 $\pm$ 0.24} &   \textbf{1.40 $\pm$ 0.23} &   \textbf{1.85 $\pm$ 0.26} &   \textbf{1.41 $\pm$ 0.23} &   \textbf{1.71 $\pm$ 0.25} \\
basic  &  \textbf{2.75 $\pm$ 0.14} &  \textbf{2.72 $\pm$ 0.14} &  \textbf{2.66 $\pm$ 0.14} &  \textbf{2.66 $\pm$ 0.14} &  \textbf{2.65 $\pm$ 0.14} &  \textbf{2.44 $\pm$ 0.14} &  2.98 $\pm$ 0.15 &  \textbf{2.57 $\pm$ 0.14} \\
rot  &  15.75 $\pm$ 0.32 &  17.09 $\pm$ 0.33 &  \textbf{14.23 $\pm$ 0.31} &  \textbf{14.98 $\pm$ 0.31} &  14.33 $\pm$ 0.31 &  15.89 $\pm$ 0.32 &  15.05 $\pm$ 0.31 &  \textbf{13.25 $\pm$ 0.30} \\
bg-img  &  20.69 $\pm$ 0.36 &  \textbf{17.01 $\pm$ 0.33} &  \textbf{17.23 $\pm$ 0.33} &  \textbf{17.44 $\pm$ 0.33} &  21.36 $\pm$ 0.36 &  \textbf{17.75 $\pm$ 0.33} &  21.77 $\pm$ 0.36 &  19.00 $\pm$ 0.34 \\
bg-rand  &  12.72 $\pm$ 0.29 &  13.65 $\pm$ 0.30 &  \textbf{8.52 $\pm$ 0.24} &  9.51 $\pm$ 0.26 &  10.72 $\pm$ 0.27 &  12.11 $\pm$ 0.29 &  \textbf{8.03 $\pm$ 0.24} &  8.70 $\pm$ 0.25 \\
bg-img-rot  &  52.44 $\pm$ 0.44 &  52.92 $\pm$ 0.44 &  \textbf{49.61 $\pm$ 0.44} &  50.93 $\pm$ 0.44 &  56.19 $\pm$ 0.43 &  56.61 $\pm$ 0.43 &  57.04 $\pm$ 0.43 &  \textbf{52.49 $\pm$ 0.44} \\
rect  &  1.14 $\pm$ 0.09 &  \textbf{0.97 $\pm$ 0.09} &  \textbf{0.83 $\pm$ 0.08} &  \textbf{0.87 $\pm$ 0.08} &  1.30 $\pm$ 0.10 &  1.85 $\pm$ 0.12 &  \textbf{0.98 $\pm$ 0.09} &  \textbf{1.26 $\pm$ 0.10} \\
rect-img  &  22.84 $\pm$ 0.37 &  22.81 $\pm$ 0.37 &  \textbf{21.96 $\pm$ 0.36} &  \textbf{21.98 $\pm$ 0.36} &  24.22 $\pm$ 0.38 &  23.46 $\pm$ 0.37 &  23.04 $\pm$ 0.37 &  \textbf{21.86 $\pm$ 0.36} \\
convex  &  28.65 $\pm$ 0.40 &  25.78 $\pm$ 0.38 &  \textbf{22.18 $\pm$ 0.36} &  25.65 $\pm$ 0.38 &  27.24 $\pm$ 0.39 &  25.00 $\pm$ 0.38 &  \textbf{20.86 $\pm$ 0.36} &  24.42 $\pm$ 0.38 \\
\bottomrule

\end{tabular}
\end{table*}

\begin{table*}[!t!b!p!h]
\caption{Linear SVM classification performance on top-level representations learnt unsupervised by deep contractive autoencoders trained from various schemes. Notations are the same as in table \ref{tbl:dae_unsup}. See text for details.}
\label{tbl:cae_unsup}
\centering
\begin{tabular}{l|cccc|cccc}
\toprule
Dataset/Method  &  CAE-2L  &  CAE-2U  &  CAE-2J  &  CAE-2UJ  &  CAE-3L  &  CAE-3U  &  CAE-3J  &  CAE-3UJ  \\
\midrule
MNIST  &  \textbf{1.55 $\pm$ 0.24} &  1.91 $\pm$ 0.27 &  \textbf{1.33 $\pm$ 0.22} &  \textbf{1.65 $\pm$ 0.25} &  \textbf{1.47 $\pm$ 0.24} &  \textbf{1.98 $\pm$ 0.27} &  \textbf{1.55 $\pm$ 0.24} &  \textbf{1.74 $\pm$ 0.26} \\
basic  &  2.96 $\pm$ 0.15 &  2.80 $\pm$ 0.14 &  \textbf{2.38 $\pm$ 0.13} &  2.70 $\pm$ 0.14 &  \textbf{2.90 $\pm$ 0.15} &  \textbf{2.92 $\pm$ 0.15} &  \textbf{2.80 $\pm$ 0.14} &  \textbf{2.95 $\pm$ 0.15} \\
rot  &  15.10 $\pm$ 0.31 &  15.29 $\pm$ 0.32 &  \textbf{13.17 $\pm$ 0.30} &  15.38 $\pm$ 0.32 &  \textbf{15.23 $\pm$ 0.31} &  16.23 $\pm$ 0.32 &  \textbf{15.14 $\pm$ 0.31} &  \textbf{15.59 $\pm$ 0.32} \\
bg-img  &  19.80 $\pm$ 0.35 &  \textbf{17.48 $\pm$ 0.33} &  19.05 $\pm$ 0.34 &  19.63 $\pm$ 0.35 &  21.31 $\pm$ 0.36 &  \textbf{19.65 $\pm$ 0.35} &  21.68 $\pm$ 0.36 &  \textbf{19.49 $\pm$ 0.35} \\
bg-rand  &  15.00 $\pm$ 0.31 &  \textbf{12.50 $\pm$ 0.29} &  14.00 $\pm$ 0.30 &  13.18 $\pm$ 0.30 &  14.30 $\pm$ 0.31 &  \textbf{12.05 $\pm$ 0.29} &  \textbf{11.53 $\pm$ 0.28} &  12.68 $\pm$ 0.29 \\
bg-img-rot  &  \textbf{52.57 $\pm$ 0.44} &  53.92 $\pm$ 0.44 &  \textbf{52.00 $\pm$ 0.44} &  \textbf{52.29 $\pm$ 0.44} &  \textbf{53.66 $\pm$ 0.44} &  55.61 $\pm$ 0.44 &  \textbf{54.32 $\pm$ 0.44} &  \textbf{53.57 $\pm$ 0.44} \\
rect  &  1.62 $\pm$ 0.11 &  1.77 $\pm$ 0.12 &  \textbf{1.07 $\pm$ 0.09} &  1.85 $\pm$ 0.12 &  1.36 $\pm$ 0.10 &  2.40 $\pm$ 0.13 &  \textbf{0.93 $\pm$ 0.08} &  1.99 $\pm$ 0.12 \\
rect-img  &  \textbf{22.62 $\pm$ 0.37} &  \textbf{22.40 $\pm$ 0.37} &  \textbf{22.23 $\pm$ 0.36} &  \textbf{22.76 $\pm$ 0.37} &  \textbf{22.37 $\pm$ 0.37} &  \textbf{22.56 $\pm$ 0.37} &  \textbf{23.06 $\pm$ 0.37} &  \textbf{22.53 $\pm$ 0.37} \\
convex  &  27.45 $\pm$ 0.39 &  25.90 $\pm$ 0.38 &  \textbf{24.91 $\pm$ 0.38} &  \textbf{25.31 $\pm$ 0.38} &  28.49 $\pm$ 0.40 &  26.66 $\pm$ 0.39 &  \textbf{24.66 $\pm$ 0.38} &  \textbf{25.26 $\pm$ 0.38} \\
\bottomrule
\end{tabular}
\end{table*}

Apart from performing joint training directly with random initialization as in equation \ref{eqn:joint_obj}, it is also possible to first apply layerwise training and then jointly train with the pre-trained weights.  Therefore, to investigate whether joint training is beneficial in this situation, we performed another set of experiments by initializing the weights of deep autoencoders using the pre-trained weights from denoising and contractive autoencoders, and further performed unsupervised joint training for 300 epochs, with and without their corresponding regularizations.  The results are shown in Tables \ref{tbl:dae_unsup} and \ref{tbl:cae_unsup}. We use a suffix `U' to indicate the results from this training procedure when the following joint training is preformed without any regularization, and `UJ' is used to indicate the case where the further joint training is performed with the corresponding regularization.   Results from `U' scheme are similar to the layerwise case, whereas, the performance from `UJ' is clearly better as compare to the layerwise case.  The above result also suggest that the performance improvement from joint training is more significant while combined with corresponding regularizations, irrespective of the parameter initialization.

In summary, the representation learned through unsupervised joint training is better as compared to the layerwise case.  In addition, it is important to apply appropriate regularizations during joint training in order to get the most benefit.



%
\subsection{How Important is the Regularization?}
From the previous experiments, it is clear that the joint training scheme has several advantages over the layerwise training scheme, and it also suggest that the use of proper regularizations is important (see the performance between `U' and `UJ' in Table \ref{tbl:dae_unsup} and \ref{tbl:cae_unsup} for example).  However, the role of the regularization for training deep autoencoder is still unclear - \emph{is it possible to achieve good performance by applying joint training without any regularization?}  To investigate this we trained two- and three-layered deep autoencoders for 300 epochs with L2 constraints on the weights\footnote{Training without any regularization resulted in very poor performance.}, for fair comparison with the previous experiments.  We again trained a linear SVM using the top-layer representation and reported the results in Table \ref{tbl:ae_unsup}.  Performance was significantly worse as compared to the case where more powerful regularizers from autoencoders were employed (see Tables \ref{tbl:dae_unsup} and \ref{tbl:cae_unsup}), especially in the case where noise was presented in the dataset.  Hence, the results strongly suggest that for unsupervised feature learning, more powerful regularization is required to achieve superior performance.  In addition, it is more beneficial to incorporate those powerful regularizations, such as denoising or contractive, during joint training (see Table \ref{tbl:dae_unsup} and \ref{tbl:cae_unsup}).
\begin{table}[!h!t!b!p]
\caption{Linear SVM classification performance on top-level representations learnt unsupervised by plain deep autoencoders with L2 weight regularization. See text for details.}
\label{tbl:ae_unsup}
\centering
\begin{tabular}{lrr}
\toprule
Dataset/Method  & \multicolumn{1}{c}{AE-2} & \multicolumn{1}{c}{AE-3} \\
\midrule
MNIST & 1.96 $\pm$ 0.27 & 2.36 $\pm$ 0.30 \\
basic & 3.20 $\pm$ 0.15 & 3.20 $\pm$ 0.15 \\
rot  &  18.75 $\pm$ 0.34 &  66.70 $\pm$ 0.41\\
bg-img  &  26.71 $\pm$ 0.39 &  86.30 $\pm$ 0.30\\
bg-rand  &  84.24 $\pm$ 0.32 &  85.95 $\pm$ 0.30\\
bg-img-rot  &  58.33 $\pm$ 0.43 &  81.00 $\pm$ 0.34 \\
rect  &  4.00 $\pm$ 0.17 &  4.15 $\pm$ 0.17\\
rect-img  &  24.30 $\pm$ 0.38 & 48.70 $\pm$ 0.44\\
convex & 43.70 $\pm$ 0.43 & 45.08 $\pm$ 0.44\\
\bottomrule
\end{tabular}
\end{table}

%
%
\begin{table*}[!tbph]
\caption{Classification performance after finetuning on deep denoising autoencoders trained with various schemes.  Notations are the same as in Table \ref{tbl:dae_unsup}.  See text for details.}
\label{tbl:dae_sup}
\centering
\begin{tabular}{l|cccc|cccc}
\toprule
Dataset/Method  &  DAE-2L  &  DAE-2U  &  DAE-2J  &  DAE-2UJ  &  DAE-3L  &  DAE-3U  &  DAE-3J  &  DAE-3UJ  \\
\midrule
MNIST  &  \textbf{1.05 $\pm$ 0.20} &  \textbf{1.17 $\pm$ 0.21} &  \textbf{1.17 $\pm$ 0.21} &  \textbf{1.21 $\pm$ 0.21} &  \textbf{1.20 $\pm$ 0.21} &  \textbf{1.15 $\pm$ 0.21} &  \textbf{1.10 $\pm$ 0.21} &  \textbf{1.26 $\pm$ 0.22} \\
basic  &  \textbf{2.55 $\pm$ 0.14} &  \textbf{2.42 $\pm$ 0.13} &  \textbf{2.44 $\pm$ 0.14} &  \textbf{2.68 $\pm$ 0.14} &  3.14 $\pm$ 0.15 &  \textbf{2.50 $\pm$ 0.14} &  \textbf{2.65 $\pm$ 0.14} &  \textbf{2.70 $\pm$ 0.14} \\
rot  &  9.51 $\pm$ 0.26 &  9.77 $\pm$ 0.26 &  \textbf{8.40 $\pm$ 0.24} &  9.57 $\pm$ 0.26 &  9.65 $\pm$ 0.26 &  9.32 $\pm$ 0.25 &  \textbf{7.87 $\pm$ 0.24} &  8.96 $\pm$ 0.25 \\
bg-img  &  \textbf{15.21 $\pm$ 0.31} &  17.08 $\pm$ 0.33 &  17.11 $\pm$ 0.33 &  17.50 $\pm$ 0.33 &  24.01 $\pm$ 0.37 &  \textbf{15.47 $\pm$ 0.32} &  18.65 $\pm$ 0.34 &  17.02 $\pm$ 0.33 \\
bg-rand  &  12.16 $\pm$ 0.29 &  11.91 $\pm$ 0.28 &  \textbf{8.98 $\pm$ 0.25} &  10.88 $\pm$ 0.27 &  18.11 $\pm$ 0.34 &  11.64 $\pm$ 0.28 &  \textbf{8.04 $\pm$ 0.24} &  \textbf{8.48 $\pm$ 0.24} \\
bg-img-rot  &  \textbf{46.35 $\pm$ 0.44} &  47.90 $\pm$ 0.44 &  \textbf{46.94 $\pm$ 0.44} &  47.71 $\pm$ 0.44 &  56.69 $\pm$ 0.43 &  \textbf{45.16 $\pm$ 0.44} &  46.95 $\pm$ 0.44 &  46.51 $\pm$ 0.44 \\
rect  &  1.60 $\pm$ 0.11 &  1.45 $\pm$ 0.10 &  \textbf{0.98 $\pm$ 0.09} &  \textbf{0.98 $\pm$ 0.09} &  1.39 $\pm$ 0.10 &  1.89 $\pm$ 0.12 &  \textbf{0.92 $\pm$ 0.08} &  \textbf{0.82 $\pm$ 0.08} \\
rect-img  &  21.87 $\pm$ 0.36 &  \textbf{21.09 $\pm$ 0.36} &  21.87 $\pm$ 0.36 &  22.53 $\pm$ 0.37 &  24.79 $\pm$ 0.38 &  \textbf{22.17 $\pm$ 0.36} &  \textbf{22.49 $\pm$ 0.37} &  \textbf{22.48 $\pm$ 0.37} \\
convex  &  \textbf{19.33 $\pm$ 0.35} &  \textbf{19.24 $\pm$ 0.35} &  \textbf{18.60 $\pm$ 0.34} &  19.88 $\pm$ 0.35 &  23.17 $\pm$ 0.37 &  \textbf{18.03 $\pm$ 0.34} &  \textbf{18.33 $\pm$ 0.34} &  19.20 $\pm$ 0.35 \\
\bottomrule
\end{tabular}
\end{table*}

\begin{table*}[!tbph]
\caption{Classification performance after finetuning on deep contractive autoencoders trained with various schemes.  Notations are the same as in Table \ref{tbl:dae_unsup}. See text for details.}
\label{tbl:cae_sup}
\centering
\begin{tabular}{l|rrrr|rrrr}
\toprule
Dataset/Method  &  CAE-2L  &  CAE-2U  &  CAE-2J  &  CAE-2UJ  &  CAE-3L  &  CAE-3U  &  CAE-3J  &  CAE-3UJ  \\
\midrule
MNIST  &  \textbf{1.25 $\pm$ 0.22} &  \textbf{1.35 $\pm$ 0.23} &  \textbf{1.09 $\pm$ 0.20} &  \textbf{1.26 $\pm$ 0.22} &  \textbf{1.47 $\pm$ 0.24} &  \textbf{1.53 $\pm$ 0.24} &  \textbf{1.11 $\pm$ 0.21} &  \textbf{1.24 $\pm$ 0.22} \\
basic  &  \textbf{2.74 $\pm$ 0.14} &  \textbf{2.85 $\pm$ 0.15} &  \textbf{2.63 $\pm$ 0.14} &  \textbf{2.74 $\pm$ 0.14} &  3.23 $\pm$ 0.15 &  \textbf{3.01 $\pm$ 0.15} &  \textbf{2.78 $\pm$ 0.14} &  \textbf{2.84 $\pm$ 0.15} \\
rot  &  10.31 $\pm$ 0.27 &  10.47 $\pm$ 0.27 &  \textbf{8.56 $\pm$ 0.25} &  10.18 $\pm$ 0.27 &  10.85 $\pm$ 0.27 &  10.04 $\pm$ 0.26 &  \textbf{7.91 $\pm$ 0.24} &  9.36 $\pm$ 0.26 \\
bg-img  &  \textbf{15.82 $\pm$ 0.32} &  18.45 $\pm$ 0.34 &  18.34 $\pm$ 0.34 &  18.45 $\pm$ 0.34 &  25.32 $\pm$ 0.38 &  \textbf{16.58 $\pm$ 0.33} &  \textbf{17.24 $\pm$ 0.33} &  \textbf{17.01 $\pm$ 0.33} \\
bg-rand  &  13.12 $\pm$ 0.30 &  \textbf{10.53 $\pm$ 0.27} &  12.99 $\pm$ 0.29 &  12.53 $\pm$ 0.29 &  18.21 $\pm$ 0.34 &  \textbf{11.19 $\pm$ 0.28} &  13.67 $\pm$ 0.30 &  12.25 $\pm$ 0.29 \\
bg-img-rot  &  \textbf{46.37 $\pm$ 0.44} &  47.38 $\pm$ 0.44 &  47.77 $\pm$ 0.44 &  48.00 $\pm$ 0.44 &  58.93 $\pm$ 0.43 &  \textbf{46.80 $\pm$ 0.44} &  \textbf{47.19 $\pm$ 0.44} &  \textbf{47.50 $\pm$ 0.44} \\
rect  &  2.17 $\pm$ 0.13 &  1.99 $\pm$ 0.12 &  \textbf{1.17 $\pm$ 0.09} &  1.39 $\pm$ 0.10 &  2.44 $\pm$ 0.14 &  2.15 $\pm$ 0.13 &  \textbf{1.04 $\pm$ 0.09} &  1.51 $\pm$ 0.11 \\
rect-img  &  \textbf{21.83 $\pm$ 0.36} &  \textbf{21.72 $\pm$ 0.36} &  \textbf{22.13 $\pm$ 0.36} &  \textbf{22.22 $\pm$ 0.36} &  24.56 $\pm$ 0.38 &  \textbf{22.03 $\pm$ 0.36} &  23.22 $\pm$ 0.37 &  23.58 $\pm$ 0.37 \\
convex  &  \textbf{18.60 $\pm$ 0.34} &  \textbf{18.63 $\pm$ 0.34} &  \textbf{18.01 $\pm$ 0.34} &  19.23 $\pm$ 0.35 &  23.20 $\pm$ 0.37 &  \textbf{18.60 $\pm$ 0.34} &  \textbf{18.00 $\pm$ 0.34} &  19.20 $\pm$ 0.35 \\
\bottomrule
\end{tabular}
\end{table*}

\subsection{How does Joint Training Affect Finetuning?}
So far, we have compared joint with layerwise training in unsupervised representation learning settings.  Now we turn our attention to supervised setting and investigate how joint training affects finetuning.  In this experiment, the unsupervised deep autoencoders were used to initialize the parameters of a multi-layer perceptron for the supervised finetuning (the same way as one would use layerwise for supervised tasks).  The finetuning was performed for all previously trained models for a maximum 1,000 epochs with early stopping on the validation set error.  As expected, the performance of the standard deep autoencoder (see Table \ref{tbl:ae_sup}) was not very impressive except on MNIST  which contained `cleaner' samples and significantly more training examples.  It is also reasonable to expect similar performance from layerwise and joint training since the supervised finetuning process adjusts all parameters to better fit $p(\vect{y|\vect{x}})$.  This is partially true as can be observed from the results in Table \ref{tbl:dae_sup} and \ref{tbl:cae_sup}.  The performance of 2-layer models are close in almost all cases.  However, in 3-layer case the models trained with joint training appear to perform better.  This is true for the models pre-trained via joint training with or without regularization (\ie scheme `U' and `UJ' respectively), which might suggests that joint training is more beneficial for deeper models.  Hence, even in the case, where one would tuning all parameters of the model for supervised tasks, unsupervised joint training can still be beneficial, especially for deeper models.  The results also suggest that as long as appropriate regularization is employed in joint pre-training, initialization does not influence the supervised performance significantly.

\begin{table}[!htb]
\caption{Classification performance of the plain deep autoencoder with L2 weight regularization after supervised finetuning.  See text for details.}
\label{tbl:ae_sup}
\centering
\begin{tabular}{lrr}
\toprule
Dataset/Method  & \multicolumn{1}{c}{AE-2} & \multicolumn{1}{c}{AE-3} \\
\midrule
MNIST & 1.36 $\pm$ 0.23 & 1.30 $\pm$ 0.22\\
basic & 3.18 $\pm$ 0.15 & 3.04 $\pm$ 0.15\\
rot  &  9.95 $\pm$ 0.26 &  9.43 $\pm$ 0.26\\
bg-img  &  24.38 $\pm$ 0.38 &  24.00 $\pm$ 0.37\\
bg-rand  &  14.47 $\pm$ 0.31 &  18.53 $\pm$ 0.34\\
bg-img-rot  &  52.00 $\pm$ 0.44 &  54.91 $\pm$ 0.44\\
rect  &  2.91 $\pm$ 0.15 &  3.08 $\pm$ 0.15\\
rect-img  &  23.45 $\pm$ 0.37 & 24.00 $\pm$ 0.37\\
convex & 21.27 $\pm$ 0.36 & 22.25 $\pm$ 0.36\\
\bottomrule
\end{tabular}
\end{table}

\section {Conclusion}
\label{sec:conclusion}

In this paper we presented an unsupervised method for jointly training all layers of deep autoencoder and analysed its performance against greedy layerwise training in various circumstances.  We formulated a single objective for the deep autoencoder, which consists of a global reconstruction objective with local constraints on the hidden layers, so that all layers could be trained jointly.  This could also be viewed as a generalization of training single- to multi-layer autoencoders, and provided a straightforward way to stack the different variants of autoencoders.  Empirically, we showed that the joint training method not only learned better data models, but also learned more representative features for classification as compared to the layerwise method, which highlights its potential for unsupervised feature learning.  In addition, the experiments also showed that the success of the joint training technique is dependent on the more powerful regularizations proposed in the more recent variants of autoencoders.  In the supervised setting, joint training also shows superior performance when training deeper models.  Going forward, this framework of jointly training deep autoencoders can provide a platform for investigating more efficient usage of different types of regularizers, especially in light of the growing volumes of available unlabeled data.


\appendix

\subsection{Additional Qualitative Samples}

\begin{figure*}[!hptb]
\centering
\begin{tabular}{cc}
\includegraphics[width=0.5\textwidth]{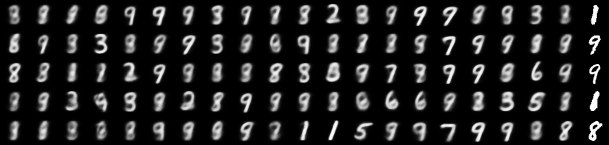} &  
\includegraphics[width=0.5\textwidth]{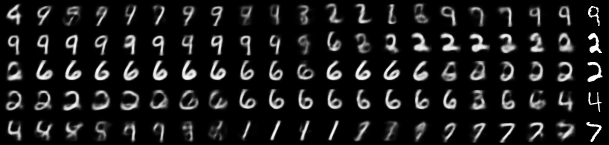} \\
\includegraphics[width=0.5\textwidth]{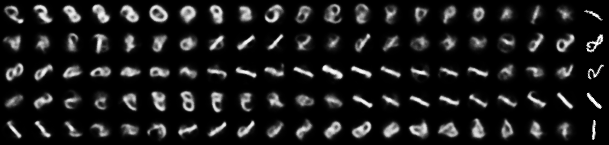} & 
\includegraphics[width=0.5\textwidth]{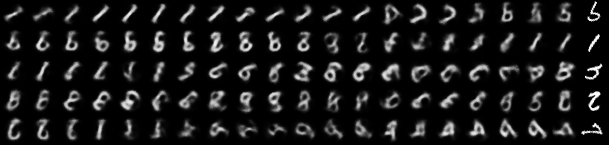} \\
\includegraphics[width=0.5\textwidth]{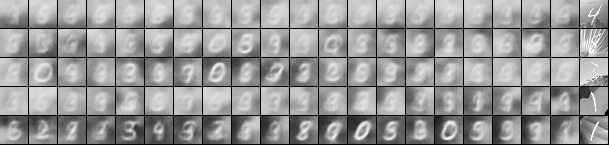} & 
\includegraphics[width=0.5\textwidth]{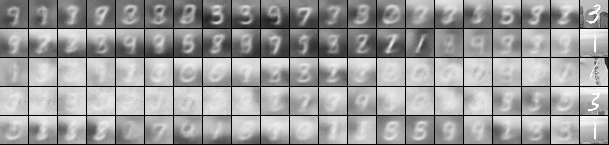} \\
\includegraphics[width=0.5\textwidth]{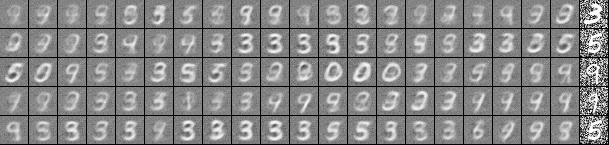} & 
\includegraphics[width=0.5\textwidth]{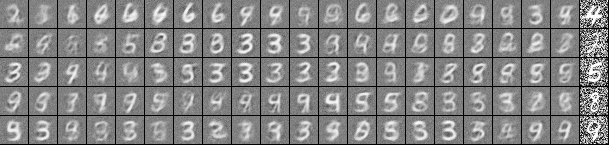} \\
\includegraphics[width=0.5\textwidth]{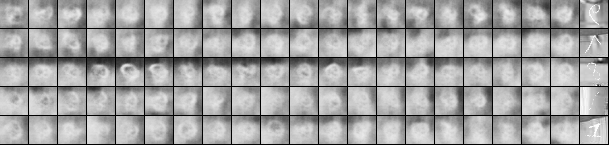} &
\includegraphics[width=0.5\textwidth]{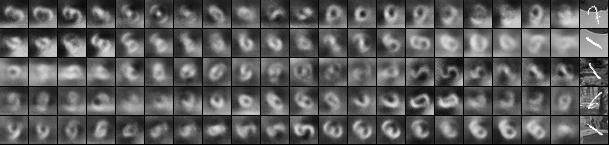} \\
\includegraphics[width=0.5\textwidth]{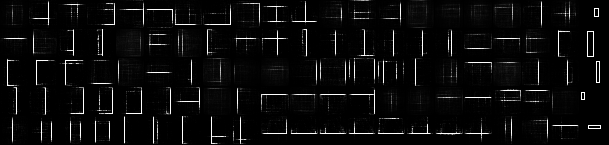} & 
\includegraphics[width=0.5\textwidth]{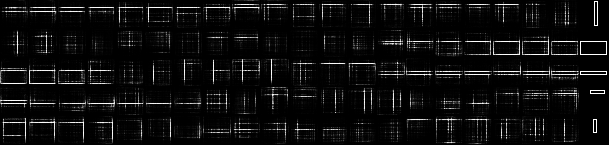} \\
\includegraphics[width=0.5\textwidth]{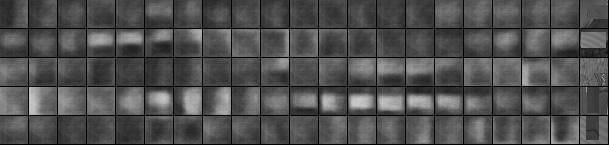} & 
\includegraphics[width=0.5\textwidth]{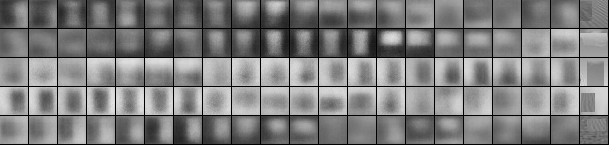} \\
\includegraphics[width=0.5\textwidth]{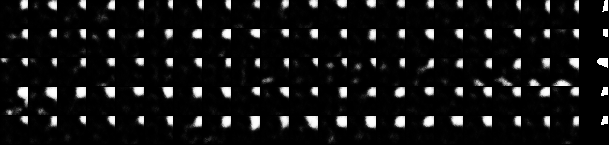} & 
\includegraphics[width=0.5\textwidth]{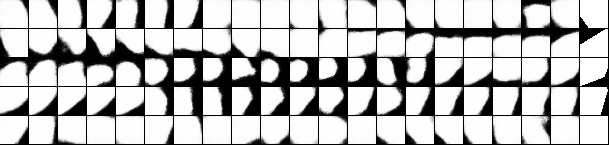} \\
\end{tabular}
\caption{Expanded samples from those are shown in Figure \ref{fig:samples} in which we show every fourth sample but here we show every consecutive sample (from left to right, top to bottom) and with longer runs.  The last column shows the closest sample from the training set to illustrate the model is not memorizing the training data.  \textbf{From top to bottom:} samples from MNIST, MNIST-rotation, MNIST-bg-image, MNIST-bg-random, MNIST-bg-rot-image, rectangle, rectangle-image and convex dataset.  \textbf{Left:} consecutive samples generated from deep denoising autoencoder trained by using layerwise scheme.  \textbf{Right:} consecutive samples generated from deep denoising autoencoder trained by joint training.  The joint trained models show sharper and more diverse samples in general.}
\label{fig:expanded_samples}
\end{figure*}

\bibliographystyle{IEEEtran}
\bibliography{main}
\end{document}